\documentclass[letterpaper]{article} 
\usepackage{aaai25}  
\usepackage{times}  
\usepackage{helvet}  
\usepackage{courier}  
\usepackage[hyphens]{url}  
\usepackage{graphicx} 
\usepackage{natbib}  
\usepackage{caption} 
\usepackage{algorithm}
\usepackage{algorithmic}
\usepackage{amsfonts}
\usepackage{amsmath}
\usepackage{multirow}

\usepackage{subcaption}
\usepackage{pdfpages}

\usepackage{times}
\usepackage{helvet}
\usepackage{courier}
\usepackage{xcolor}
\nocopyright
\usepackage{newfloat}
\usepackage{listings}
\DeclareCaptionStyle{ruled}{labelfont=normalfont,labelsep=colon,strut=off} 
\floatstyle{ruled}
\newfloat{listing}{tb}{lst}{}
\floatname{listing}{Listing}
\title{Geo2Vec: Shape- and Distance-Aware Neural Representation of Geospatial Entities}
\author{
    Chen Chu,
    Cyrus Shahabi\\
}
\affiliations{
    University of Southern California\\
    Los Angeles, CA USA\\
    chenchu@usc.edu, shahabi@usc.edu
}

\usepackage{bibentry}

\begin{document}

\maketitle

\begin{abstract}
Spatial representation learning is essential for GeoAI applications such as urban analytics, enabling the encoding of shapes, locations, and spatial relationships (topological and distance-based) of geo-entities like points, polylines, and polygons.
Existing methods either target a single geo-entity type or, like Poly2Vec, decompose entities into simpler components to enable Fourier transformation, introducing high computational cost. Moreover, since the transformed space lacks geometric alignment, these methods rely on uniform, non-adaptive sampling, which blurs fine-grained features like edges and boundaries.
To address these limitations, we introduce Geo2Vec, a novel method inspired by signed distance fields (SDF) that operates directly in the original space. Geo2Vec adaptively samples points and encodes their signed distances (positive outside, negative inside), capturing geometry without decomposition. A neural network trained to approximate the SDF produces compact, geometry-aware, and unified representations for all geo-entity types. Additionally, we propose a rotation-invariant positional encoding to model high-frequency spatial variations and construct a structured and robust embedding space for downstream GeoAI models. Empirical results show that Geo2Vec consistently outperforms existing methods in representing shape and location, capturing topological and distance relationships, and achieving greater efficiency in real-world GeoAI applications. Code and Data can be found at: 

\url{https://github.com/chuchen2017/GeoNeuralRepresentation}.

\end{abstract}

%

\section{Introduction}

Representation learning for geospatial entities, such as points, lines, and polygons, has become crucial for deep neural network models aiming to effectively address various downstream geospatial tasks. The ability to learn robust and unified embeddings for these entities facilitates generalization across diverse GeoAI applications, including land use classification \cite{regiondcl}, population prediction \cite{population}, urban flow inference \cite{cityfundation}, and urban morphology analysis \cite{morphology}. 

Several Spatial Representation Learning (SRL) approaches have been developed specifically for individual entity types like lines and polygons. For example, polyline-based methods often employ sequence models like RNN or Transformer \cite{TJEPA}, but these approaches primarily capture vertex connectivity and largely overlook crucial geometric and topological details associated with line segments. Similarly, polygon-specific methods typically use graph neural networks (GNNs) to represent vertices and edges as graph components \cite{PolyGNN}, yet these methods inadequately preserve the polygons' spatial extent (interior and exterior) and often struggle with complex geometries, particularly polygons with holes.

\begin{figure}[h]
    \centering
    \includegraphics[width=\linewidth]{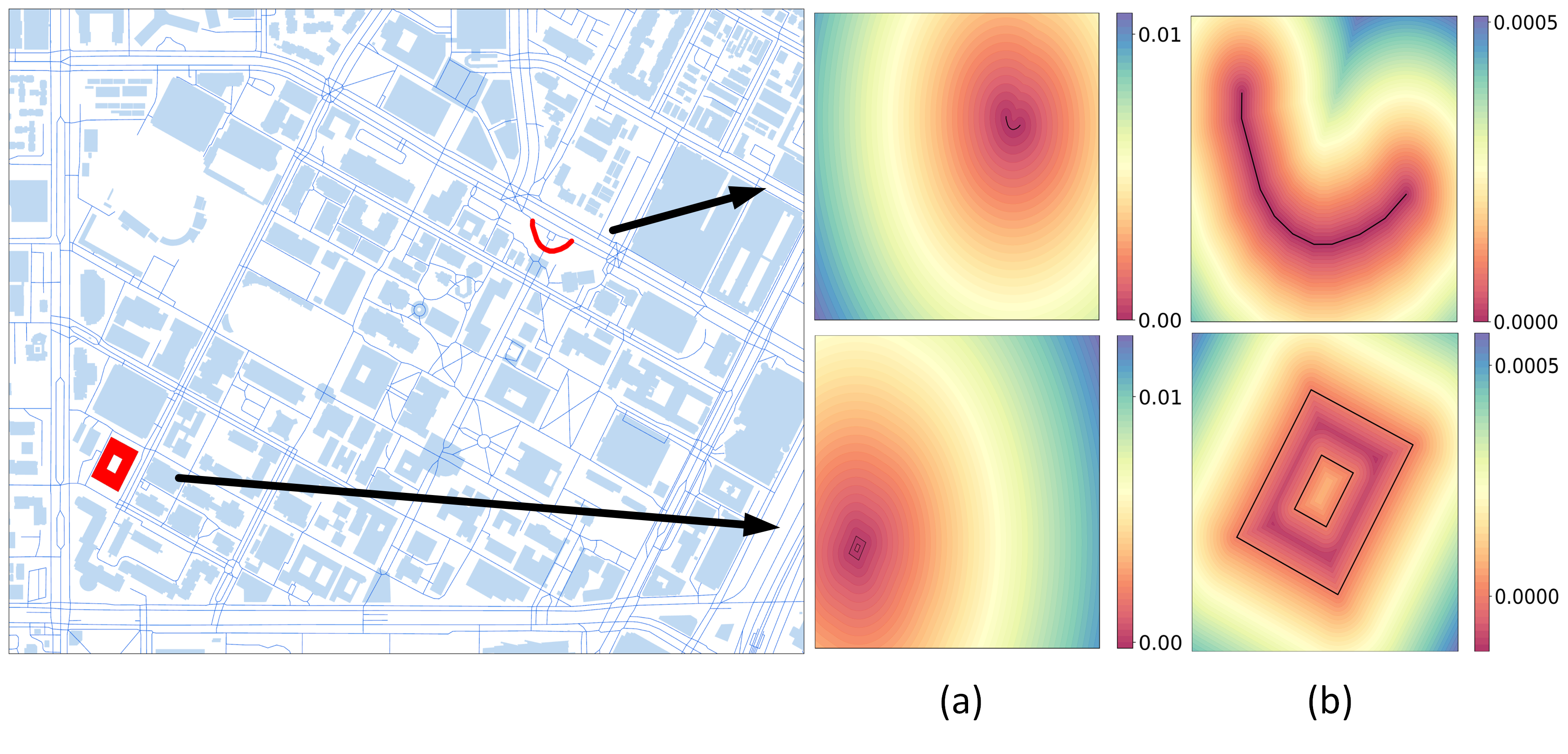}
    \caption{Signed Distance Fields for two types of geo-entities at spatial scales: (a) coarse scale, (b) fine scale.}
    \label{Figure 1}
\end{figure}

To enable unified embeddings across all geo-entity types, recent methods like Poly2Vec \cite{poly2vec} decompose complex entities into simpler components suitable for Fourier transformation. This decomposition, combined with the computational overhead of performing the Fourier transformation itself, results in high processing cost. Moreover, since the transformed Fourier space lacks direct correspondence with the original geometry and topology, these methods are limited to uniform, non-adaptive sampling, which fails to preserve fine-grained geometric features like edges and boundaries.

Consequently, there is a need for a unified SRL approach that captures geometry and location across all geospatial entity types while performing well on standard evaluation tasks like capturing topological and distance relationships \cite{Ji2025}, to ensure effectiveness in real-world GeoAI tasks.

Towards this end, we propose Geo2Vec, a neural representation approach that explicitly learns a Signed Distance Field (SDF) of each geospatial entity. Specifically, the SDF is defined as the shortest distance from any point in space to the boundary of the entity, with negative values indicating points inside the entity and positive values outside. Examples of SDFs for polygon- and polyline-type geo-entities are shown in Figure~\ref{Figure 1}. In Figure~\ref{Figure 1}(a), the coarse-scale SDF clearly captures the spatial locations of the two geo-entities, with low-value regions highlighted in red. In contrast, the fine-scale SDFs in Figure~\ref{Figure 1}(b) capture the detailed shapes of the entities as continuous fields. Notably, Geo2Vec’s use of SDFs enables a \textbf{unified} representation across all geo-entity types: polygons with spatial extent (including those with holes) yield negative SDF values within their interiors, while points and lines, lacking interior regions, do not. This continuous and differentiable representation overcomes all the limitations of discrete vertex-edge models and entity-specific decompositions.

Moreover, Geo2Vec leverages coordinates in the original (non-transformed) space, allowing strategic sampling near entity boundaries or regions requiring higher precision. This adaptive sampling significantly enhances representation quality. Notably, our experiments show that our adaptive sampling method significantly outperform Fourier space sampling, achieving comparable accuracy with less than 35\% of samples and thus delivering superior efficiency.

Our empirical evaluations show that Geo2Vec significantly outperforms SOTA methods on standard evaluation tasks for shape and location representation, achieving improvement up to 61.95\% and 54.3\%, respectively. Finally, we introduce a rotation-invariant positional encoding that produces a more structured and robust embedding space, where geo-entities with similar shapes are positioned closer together regardless of their orientation. This property is useful for unsupervised models and supervised models with weaker learning signals, and our experiments specifically demonstrate its effectiveness in improving Geo2Vec performance in unsupervised downstream tasks.

\section{Related Works}

Spatial Representation Learning (SRL) aims at directly learning the neural representation of various types of spatial data in their native format without the need for feature engineering and data conversion stage \cite{SRL}. Most prior work has focused on learning representations for different types of spatial data in isolation. For example, point encoding \cite{mai2023CSP}, trajectory representation learning \cite{Jingyuan2023}, road network representation \cite{Liang2025}, and polygon representation \cite{contrastive2024}. Current methods primarily rely on feeding the discrete data structures directly into neural models to learn representations \cite{Traj2024,PolyGNN}. Although, the discrete vertex/edge representation is suitable for data storage and visualization clarity, it is not effective for representing spatial extent or their topological characteristics. This mismatch between representation format and geo-entity leads to limitations in expressiveness.

The current state-of-the-art method, Poly2Vec \cite{poly2vec}, encodes points, polylines, and polygons using Fourier transforms. However, existing research has not revealed the relationship between real-world coordinates and the Fourier feature space (Spectral domain). As a result, Fourier-based methods typically employ non-adaptive, heuristic sampling strategies. Although some approaches incorporate geometric frequency selection and improve feature expressiveness \cite{NUFT}, they still fall short of identifying the most discriminative frequencies. This limitation fundamentally restricts the representational power of Fourier-based encoding, particularly when sampling frequencies are low. Moreover, applying Fourier transforms to complex objects like polygons is nontrivial, which is why they must first be decomposed into simpler shapes like triangles, further adding to the already high computational cost of the transformation.

Employing neural networks to learn continuous fields is a widely studied topic in 3D computer vision. Prior work has explored learning 3D shape representations through signed distance function \cite{DeepSDF,NEURIPS2024} and occupancy fields \cite{occupancy}. These field learning methods have shown strong effectiveness in modeling complex scenarios, as shown by NeRF \cite{NERF} and 3DGS \cite{3DGS}. However, most of this research focuses on accurately reconstructing specific shapes or scenes, rather than leveraging field-based representations for broader downstream geospatial tasks. In contrast, our work aims to learn a \textbf{generalizable embedding space} from SDFs, explicitly designed to efficiently capture geospatial semantics and (topological and distance) relationships.

\section{Preliminary}

\textbf{Definition 1}: \textbf{Spatial Position} refers to the location of a geo-entity expressed in geographic coordinates or a global reference system, representing its precise placement in physical or world space.\\
\textbf{Definition 2}: \textbf{Spatial Extent} refers to the coverage area of a geo-entity, representing its shape and spatial footprint.\\
\textbf{Definition 3}: A \textbf{Geo-entity} \(E\) is an object characterized by its spatial position and, optionally, its spatial extent. It is generally recorded in a sequence of coordinates \(P_{E} = \{\mathbf{x}_i\}_{i=1}^{N} \in \mathbb{R}^{N \times 2}\), where \(\mathbf{x}_i = (x_i,y_i)\) denotes a vertex, and \(N\) denotes the number of vertices. Common examples include points, polylines, polygons, multi-polygons, and other related spatial data types.\\
\textbf{Definition 4} (Signed Distance Function). Given a geo-entity \(E\) and a query point in space \(x \in \mathbb{R}^2\), the signed distance function \(\mathbf{SDF}(\mathbf{x},E)=s\) returns the shortest distance \(s \in \mathbb{R}\) from the query point \(\mathbf{x}\) to the boundary of \(E\). For any geo-entity with spatial extent, the value \(s\) is positive if \(\mathbf{x}\) lies outside the spatial extent of \(E\), and negative if \(\mathbf{x}\) lies inside.\\
\textbf{Definition 5} (Signed Distance Field). The Signed Distance Field of a geo-entity \( E \) is a scalar field defined over a continuous spatial domain \( \Omega_{E} \subseteq \mathbb{R}^2 \), in which each point \( \mathbf{x} \in \Omega_{E} \) is assigned a scalar value representing its signed distance to \( E \). This field provides a continuous representation of the geo-entity, capturing both location and shape information.\\
\textbf{Definition 6} (Representation Learning of Geo-entity). Given a dataset of geo-entities \(G = \{E_i\}_{i=1}^{N}\), the goal of representation learning is to learn a mapping function \(\mathcal{E}_\theta: E \rightarrow \mathbf{z}_{E} \in \mathbb{R}^d\), where \(\mathbf{z}_{E} \in \mathbb{R}^d\) is a \(d\)-dimensional embedding. The learned representations should preserve the data utility of the original formats, allowing effective support for a range of spatial reasoning tasks. Moreover, by unifying different types of geo-entities into a common representation embedding, the representation becomes broadly applicable across diverse downstream models.

\section{Nueral Representation of Geo-entity}

\begin{figure*}
\centering
\includegraphics[width=\linewidth]{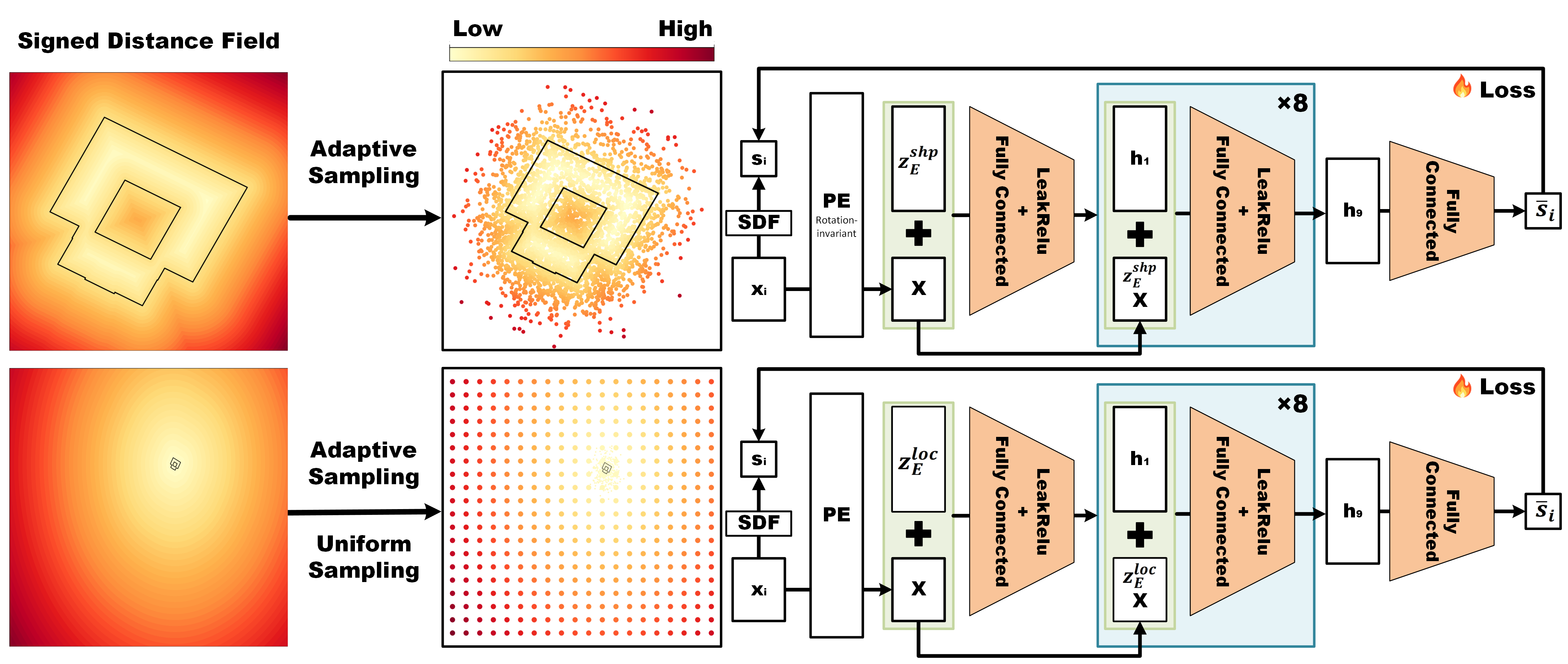}
\caption{An illustration of the Geo2Vec learning framework.} 
\label{Figure 2}
\end{figure*}

In this section, we present Geo2Vec, which aims to learn the representation of a geo-entity by explicitly modeling its SDF. We employ a neural network \(\mathcal{G}_{\theta}\) to approximate the signed distance function \(\mathbf{SDF}(\mathbf{x}, E)\) and learn the corresponding SDF \(\Omega_E\). To achieve this, for each geo-entity \(E\), we sample a set of training points \(X_E := \{(\mathbf{x}, s) \mid s = \mathbf{SDF}(\mathbf{x},E) \}\). Then train the neural network \(\mathcal{G}_{\theta}\) to learn the underlying SDF based on sample points.

For polygon shape learning, we scale each polygon individually to a canonical space $[-1,1] \times [-1,1]$ and then learn its scale-invariant shape embedding \(\mathbf{z}_E^{shp}\). For location representation learning, we normalize the entire dataset \(G\) to a canonical space and then learn the location representation of each geo-entity \(\mathbf{z}_E^{loc}\). The final representation is formed by concatenating the location and shape vectors: $\mathbf{z}_{E} = [\mathbf{z}_{E}^{loc}, \mathbf{z}_{E}^{shp}]$. For point entities, we use a uniform vector as their shape representation \(\mathbf{z}_{E}^{shp}\). The learning pipelines for shape and location representation are identical, the major differences lie in the sampling strategy and the positional encoding method. 

\subsection{An Adaptive Sampling Strategy}

One key advantage of representing a geo-entity using its SDF is that it allows us to directly sample in the coordinate space. To better leverage this property, we propose an adaptive sampling strategy that adjusts sampling parameters based on the learning objective and characteristics of the dataset. We first introduce our sampling methods, then describe how the associated parameters are tuned accordingly.

Firstly, we sample \(N_{\text{Vertex}}\) points near each vertex \(V\) of a geo-entity \(E\) following a 2D normal distribution:
\[\mathbf{x}_i\sim \mathcal{N}(P_{E},\sigma^2 I)\]
where \(\sigma\) is the standard deviation controlling the sampling radius. Each sampled point is paired with its signed distance value to construct the training set \(X_{E}^{\text{Vertex}}\):
\[X_{E}^{\text{Vertex}} = {\{\mathbf{x}_i,\mathbf{SDF}(\mathbf{x}_{i},E)\}}_{i=1}^{N_{\text{Vertex}}} \] 

To enhance boundary coverage, we introduce stochastic perpendicular sampling, which perturbs sampled points along each edge by applying a small normal-direction offset drawn from a symmetric distribution. Formally, for any two continuous points \(\mathbf{x}_i,\mathbf{x}_{i+1} \in P_E\), \(P_E\) denotes the sequence of coordinates of entity \(E\), we sample \({\mathbf{x}}^{\prime}\) according to the following formulate: 
\[{\mathbf{x}}^{\prime} = (1-f) \mathbf{x}_i+f \mathbf{x}_{i+1}+s \cdot d \cdot \frac{1}{|| {p2-p1} ||} \cdot \begin{pmatrix}
-(y_{i+1} - y_{i}) \\
x_{i+1} - x_{i}
\end{pmatrix}\]
where \(f \sim \mathcal{U}(0, 1)\) controls the position along the edge, \(d \sim \mathcal{N}(0,\sigma^2 I)\) specifies the magnitude of the perpendicular offset, and \(s \sim \mathcal{U}\{-1,+1\}\) randomly selects the side of the edge. Stochastic perpendicular sampling improves the model’s ability to capture the edge position in the SDF. Similarly, we construct the training dataset by sampling \(N_{\text{Edge}}\) points for each \(Edge\):
\[X_{E}^{\text{Edge}} = {\{\mathbf{x}_i,\mathbf{SDF}(\mathbf{x}_{i},E)\}}_{i=1}^{N_{\text{Edge}}} \] 

Lastly, we uniformly sample points from the coordinate space to capture the global structure of the geo-entity and to fill in regions that may have been overlooked by the previous two sampling stages. Specifically, we sample \(N_{\text{axis}}\) points along each axis, resulting in a total of \(N_{\text{Space}} = {N_{\text{axis}}}^2\) uniformly distributed points across the space, which constitute the dataset \(X_E^{\text{space}}\).
\[X_{E}^{\text{Space}} = {\{\mathbf{x}_i,\mathbf{SDF}(\mathbf{x}_{i},E)\}}_{i=1}^{N_{\text{Space}}} \] 

Finally, we combine all sampled points to form our training dataset: 
\[X_E=\{X_{E}^{\text{Vertex}},X_{E}^{\text{Edge}},X_{E}^{\text{Space}}\}_{\text{Vertex} \in E , \text{Edge} \in E}\]

During the sampling process, we leave several parameters flexible, allowing Geo2Vec to adaptively sample based on the data distribution of the target dataset. 

When learning the location representation, which aims to capture the spatial relationships among geo-entities, it is important to provide the model with information about its local neighborhood. Therefore, we set the sampling parameters according to the distances between geo-entities. It is worth noting that, although it would be beneficial to sample a variable number of points for different geo-entities, we fix the following parameters as global constants within each dataset to ensure computational efficiency. Specifically, we randomly sample a subset of geo-entities $E$, compute the distances to their $k$ nearest neighbors, and define the location sampling parameter $\sigma_{\text{loc}}$ as the standard deviation of the resulting distance distribution.

For learning the shape representation, we follow a similar strategy, but base it on edge distances. A subset of geo-entities $E$ is randomly selected, and for each of their edges, we compute the distances to their top $k$ nearest edges. The standard deviation of these distances is then used to define the shape sampling parameter $\sigma_{\text{shp}}$.

After determining the sampling deviation $\sigma$, we introduce a resolution parameter $\epsilon$ to decide the number of points to sample per unit. This parameter controls how finely we capture local spatial variation. The number of samples is given by: $N_{\text{Vertex}} = \pi  \sigma^2\epsilon^2$, $N_{\text{Edge}} = 2 \sigma  l_{\text{Edge}}\epsilon^2$, where $l_{\text{Edge}}$ denotes the length of the \(Edge\). For computational simplicity, we approximate them as: $N_{\text{Vertex}} = \epsilon \cdot \sigma$, $N_{\text{Edge}} = \epsilon \cdot l_{\text{Edge}}$, which retains the core dependency on resolution, neighborhood spread, and edge length. 

\subsection{Positional Encoding}
SDF shows various spatial patterns in different scales, and successfully modeling these patterns is crucial for its representation. We employ a Positional Encoding (PE) that maps the spatial coordinate \(\mathbf{x} \in \mathbb{R}\) to a higher dimensional space \( \mathbb{R}^{2L}\), providing spatial features that encode local and global signed distance variation. The positional encoding is formulated as follows:
\begin{align*}
PE(\mathbf{x}) = (&\sin(2^{L_{\min}} \pi \mathbf{x}), \cos(2^{L_{\min}} \pi \mathbf{x}), \ldots, \\
                 &\sin(2^{L_{\max}} \pi \mathbf{x}), \cos(2^{L_{\max}} \pi \mathbf{x}))
\end{align*}
where \(L_{\min}\) and \(L_{\max}\) define the lower and upper bounds of frequency levels, and we uniformly sample \(L\) frequencies in this bound. Unlike the positional encoding used in Transformer and NERF, we do not predefine these bounds. Instead, we set \(L_{\min}\) and \(L_{\max}\) based on the distribution of geo-entities and the specific learning objective.

Positional encoding is essential for both shape and location learning, but for opposite reasons. When learning location representations, the model aims to capture the coarse-scale trends of SDF, which, as shown in Figure \ref{Figure 1}(a), decreases uniformly in all directions and is largely independent of the specific shape of the entity. Thus, positional encoding helps encode such global variation and is expected to generalize across geo-entities. In this case, high-frequency components will introduce repeated features that hinder learning of smooth global patterns, so we avoid using these repeated frequencies when leaning location representation.

In contrast, shape representation learning focuses on capturing fine-grained, local variations in SDFs that are unique to each geo-entity, as shown in Figure \ref{Figure 1} (b). Modeling such fine spatial variations requires encoding the input coordinates with high-frequency signals, which enables the model to represent sharp transitions. These patterns are typically difficult for neural networks to learn from smooth coordinate inputs alone. Therefore, we have the following settings for positional encoding:
\[
\Delta_x = \max_{E \in G} (E.x) - \min_{E \in G} (E.x), 
\Delta_y = \max_{E \in G} (E.y) - \min_{E \in G} (E.y)
\]
\[
\Delta_{\min} = \min(\Delta_x, \Delta_y),
\Delta_{\max} = \max(\Delta_x, \Delta_y)
\]
\[
L_{\max}^{\text{loc}} \leq \log_2\left( \frac{2}{\Delta_{\min}} \right), \quad
L_{\max}^{\text{shp}} \geq \log_2\left( \frac{2}{\Delta_{\min}} \right)
\]
\[
L_{\min}^{\text{loc}},
L_{\min}^{\text{shp}} \leq 1 - \log_2\left({\Delta_{\max}} \right)
\]
Following the rules, \(L_{\max}\) and \(L_{\min}\) can be determined according to the dataset. We leave the number of sampling frequencies \(L\) as a hyperparameter.

When learning shape representations, it is important that the model learns similar embeddings for geo-entities with the same shape but different orientations. To achieve that, we propose a rotation-invariant positional encoding method, which can be formulated as: 
\[
PE_R(\mathbf{x}) = PE(\mathbf{x}'), \quad \text{where} \quad 
\mathbf{x}' = 
\begin{bmatrix}
x \\
y \\
r
\end{bmatrix}, 
r = \sqrt{x^2 + y^2}
\]
The method transforms each point’s Cartesian coordinates $\mathbf{x}$ into polar coordinates and augments the original input with the radial distance $r$. This augmentation introduces rotation-invariant features into the positional encoding, encouraging the model to capture shape geometry rather than absolute orientation. As a result, the learned embeddings become more structured and robust—a property that is particularly valuable for unsupervised downstream GeoAI models. 

\begin{table*}[htb]
\centering
\caption{Model accuracy on Shape Classification (\textbf{Shape}) and MAE on Predicting the Number of Edges (\textbf{Edge}). All accuracy values are scaled by \(\times10^{-2}\). In all tables, the values after $\pm$ indicate the standard-deviation, and \textbf{Best} results are highlighted.}
\setlength{\tabcolsep}{2pt} 
\label{Table 1}
\begin{tabular}{llrrrrrrrr}
\hline
\multicolumn{1}{c}{} & \multicolumn{2}{c}{Building}                         &  & \multicolumn{2}{c}{MNIST}                            &  & \multicolumn{1}{c}{Singapore} &  & \multicolumn{1}{c}{NYC}  \\ \cline{2-3} \cline{5-6} \cline{8-8} \cline{10-10} 
                     & \multicolumn{1}{c}{Shape$\uparrow$} & \multicolumn{1}{c}{Edge$\downarrow$} &  & \multicolumn{1}{c}{Shape$\uparrow$} & \multicolumn{1}{c}{Edge$\downarrow$} &  & \multicolumn{1}{c}{Edge$\downarrow$}      &  & \multicolumn{1}{c}{Edge$\downarrow$} \\ \hline
PolyGNN              & 87.84$\pm$0.005             & \multicolumn{1}{c}{--}                       &  & 7.77$\pm$0.013            & \multicolumn{1}{c}{--}                       &  & \multicolumn{1}{c}{--}                           &  & \multicolumn{1}{c}{--}                       \\
NUFTSPEC             & 90.46$\pm$0.730             & 3.04$\pm$0.125               &  & 96.90$\pm$0.116             & 16.76$\pm$0.588              &  & 3.66$\pm$0.023                    &  & 1.45$\pm$0.020               \\
Poly2Vec             & 76.59$\pm$1.403             & 3.34$\pm$0.127               &  & 92.52$\pm$0.265             & 29.29$\pm$0.807              &  & 3.68$\pm$0.109                    &  & 1.21$\pm$0.045               \\
Geo2Vec              & \textbf{97.34$\pm$0.310}    & \textbf{2.22$\pm$0.050}      &  & \textbf{97.58$\pm$0.097}   & \textbf{9.45$\pm$0.124}     &  & \textbf{1.40$\pm$0.011}           &  & \textbf{0.72$\pm$0.002}      \\ \hline
\end{tabular}
\end{table*}

\begin{table}[htb]
\setlength{\tabcolsep}{4pt} 
\centering
\caption{Model performance on Length of Line prediction, evaluated by MAE. All values are scaled by \(\times 10^{-4}\)}
\label{Table 2}
\begin{tabular}{lrlr}
\hline
         & \multicolumn{1}{c}{Singapore}   &  & \multicolumn{1}{c}{NYC}         \\ \cline{2-2} \cline{4-4} 
         & \multicolumn{1}{c}{Line Length$\downarrow$} &  & \multicolumn{1}{c}{Line Length$\downarrow$} \\ \hline
T2Vec    & 10.38$\pm$0.45                      &  & 13.20$\pm$0.42                      \\
T-JEPA   & 10.25$\pm$0.54                      &  & 12.65$\pm$0.36                      \\
Poly2Vec & 13.55$\pm$0.79                      &  & 21.11$\pm$0.48                      \\
Geo2Vec  & \textbf{5.75$\pm$0.26}                       &  & \textbf{7.07$\pm$0.16}                       \\ \hline
\end{tabular}
\end{table}
\subsection{Geo2Vec Model}

Given a set of sampled points, we propose the Geo2Vec model \(\mathcal{G}_{\theta}\) to learn the SDF of a geo-entity based on sampled points \(X_E\):
\[
\mathcal{G}_{\theta}(X_E, \mathbf{z}_E) \approx \mathbf{SDF}(X_E, E), \quad \forall \mathbf{x} \in X_E \subset \Omega_E
\]

Following \cite{DeepSDF}, we formulate this problem from a probabilistic perspective. We define the posterior distribution over the latent code \( \mathbf{z}_E \) given the sampled points \(X_E\) as:
\[
p_{\theta}(\mathbf{z}_E \mid X_E) = p(\mathbf{z}_E) \prod_{(\mathbf{x}_i, s_i) \in X_E} p_{\theta}(s_i \mid \mathbf{z}_E; \mathbf{x}_i)
\]
where \( p(\mathbf{z}_E) \) denotes the prior distribution over latent codes, which we assume to follow a multivariate Gaussian \( \mathcal{N}(\mathbf{0}, \sigma_z^2 \mathbf{I}) \), \(\sigma_z\) controls the density of the latent distribution. And the conditional likelihood \( p_{\theta}(s_i \mid \mathbf{z}_E; \mathbf{x}_i) \), can be expressed as:
\[
p_{\theta}(s_i \mid \mathbf{z}_E; \mathbf{x}_i) \propto \exp\left(-\mathcal{L}(\bar{s}_i, s_i)\right)
\]
where \( \bar{s}_i = \mathcal{G}_{\theta}(\mathbf{z}_E, \mathbf{x}_i) \) is the predicted signed distance at coordinate \( \mathbf{x}_i \), and \( \mathcal{L} \) is SDF the loss function.

Therefore, maximizing the posterior probability \( p_{\theta}(\mathbf{z}_E \mid X_E) \) is equivalent to minimizing the summed loss between the predicted and observed signed distances, along with a regularization term on the latent code. This formulation justifies that approximating the signed distance field using \( \mathcal{G}_\theta \) directly induces learning of the optimal latent representation \( \mathbf{z}_E \) for each geo-entity. Therefore, the resulting loss function for training Geo2Vec over the dataset \(G\) can be expressed as:
\[
\mathcal{L}_{\text{Geo2Vec}} = \sum_{E \in G} \left( \sum_{(\mathbf{x}_i, s_i) \in X_E} \mathcal{L}(\mathcal{G}_\theta(\mathbf{z}_E, \mathbf{x}_i), s_i) + \frac{\gamma}{\sigma_z^2} \|\mathbf{z}_E\|_2^2 \right)
\]
where \( \gamma \) is a hyperparameter that controls how strongly the latent codes are encouraged to follow the prior distribution. We set \( \gamma = 0 \) when learning location representations, as the spatial variation across geo-entities is sufficiently large. In this case, enforcing a tightly clustered latent space can negatively impact learning by suppressing the natural diversity of location information.

During the training process, to encourage the latent representations to reside in a shared and structured space, we jointly optimize the posterior over a large batch that includes as many geo-entities as possible. This joint training helps the model learn consistent and meaningful representations across different entities. The detailed representation learning algorithm is described in Algorithm \ref{Learning}. 
\begin{algorithm}[!h]
    \caption{Geo2Vec Training Algorithm}
    \label{Learning}
    
    \renewcommand{\algorithmicrequire}{\textbf{Input:}}
    \renewcommand{\algorithmicensure}{\textbf{Output:}}
    
    \begin{algorithmic}[1]
        \REQUIRE \(G=\{E\}\)
        \REQUIRE Sample Density \(\epsilon\), Uniform \({N_{\text{axis}}}\), batch size \( b \)
        \ENSURE \(\{\mathbf{z}_{E}\}_{E \in G}\)
        \STATE Initialize \(\mathcal{G}_{\theta}\), \(\{z_E\}_{E \in G}\sim \mathcal{N}(\mathbf{0}, \sigma_z^2 \mathbf{I})\)
        \STATE Initialize \(N_\text{Edge}\), \(N_{\text{Vertex}}\), \(\sigma\), \(X_G=\{\}\)
        \FOR{each \( E \in G \)}
            \STATE \( X_E \sim \text{Sample}(E, N_{\text{Edge}}, N_{\text{Vertex}},{N_{\text{axis}}}, \sigma) \)
            \STATE \( X_G \gets X_G \cup \{(\mathbf{x}_i, s_i, E) \mid (\mathbf{x}_i, s_i) \in X_E\} \) 
        \ENDFOR
        \STATE shuffle \(X_G=\{(x_i,s_i,E_i)\}\)     
        \FOR{each mini-batch \( \{(x_i, s_i, E_i)\}_{i=1}^{b} \subset X_G \)}
            \STATE \( \mathcal{L} = \mathcal{L}_{\text{Geo2Vec}}(\{(x_i, s_i, E_i)\}_{i=1}^b) \)
            \STATE Update \( \{\mathbf{z}_{E_i}\}_{E_i \in b} \) and \( \mathcal{G}_{\theta} \) using \( \mathcal{L} \)
        \ENDFOR
        \RETURN \(\{\mathbf{z}_E\}_{E \in G}\)
    \end{algorithmic}
\end{algorithm}

The architecture of the Geo2Vec network is shown in Figure~\ref{Figure 2}. Instead of using ReLU, we employ LeakyReLU as the activation function, as learning the negative interior structure is also crucial, and LeakyReLU preserves a non-zero gradient in the negative domain. The input point \(\mathbf{x}_i\) is projected by \(PE\) to \(\mathbf{X}\), which is then concatenated with the latent representation \(\mathbf{z}_E\). This combined vector is then concatenated with the hidden states of the neural network at each layer, serving as a conditioning input for prediction.

\section{Experimental Evaluation}
\begin{table*}[]
\centering
\caption{Overall model performance on distance estimation, evaluated by MAE. All values are scaled by \(\times 10^{-3}\).}
\setlength{\tabcolsep}{2pt} 
\label{Table 3}
\begin{tabular}{lrrrrrrrrrrr}
\hline
         & \multicolumn{1}{c}{Building} &  & \multicolumn{1}{c}{MNIST} &  & \multicolumn{3}{c}{Singapore}                                                     &  & \multicolumn{3}{c}{NYC}                                                           \\ \cline{2-2} \cline{4-4} \cline{6-8} \cline{10-12} 
         & \multicolumn{1}{c}{Pg-Pg$\downarrow$}    &  & \multicolumn{1}{c}{Pg-Pg$\downarrow$} &  & \multicolumn{1}{c}{Pt-Pg$\downarrow$} & \multicolumn{1}{c}{Pl-Pg$\downarrow$} & \multicolumn{1}{c}{Pg-Pg$\downarrow$} &  & \multicolumn{1}{c}{Pt-Pg$\downarrow$} & \multicolumn{1}{c}{Pl-Pg$\downarrow$} & \multicolumn{1}{c}{Pg-Pg$\downarrow$} \\ \hline
TILE     & 217.1$\pm$1.6                      &  & 223.8$\pm$1.0                   &  & 99.9$\pm$1.7                  & 115.5$\pm$1.5                   & 114.3$\pm$1.4                   &  & 127.6$\pm$0.7                 & 154.3$\pm$2.1                   & 167.4$\pm$2.0                   \\
THEORY   & 7.3$\pm$4.3                     &  & 34.2$\pm$1.0                  &  & 24.3$\pm$0.9                 & 25.0$\pm$0.4                 & 25.0$\pm$1.1                  &  & 26.2$\pm$1.1                 & 26.6$\pm$0.5                 & 27.4$\pm$0.7                 \\
Poly2Vec & 13.1$\pm$1.0                     &  & 21.0$\pm$0.4                 &  & 15.9$\pm$0.6                 & 19.9$\pm$1.7                  & 22.0$\pm$0.6                 &  & 28.7$\pm$1.4                 & 28.5$\pm$0.4                 & 52.7$\pm$0.8                 \\
Geo2Vec  & \textbf{6.4$\pm$0.9}                     &  & \textbf{13.0$\pm$0.8}                  &  & \textbf{5.4$\pm$0.5}                 & \textbf{5.0$\pm$0.1}                & \textbf{5.5$\pm$0.4}                 &  & \textbf{10.2$\pm$0.1}                & \textbf{13.0$\pm$0.9}                 & \textbf{12.9$\pm$0.6}                 \\ \hline
\end{tabular}
\end{table*}

\begin{table*}[]
\centering
\setlength{\tabcolsep}{2pt} 
\caption{Model accuracy on Topological Relationship Classification. All values are scaled by \(\times 10^{-2}\).}
\label{Table 4}
\begin{tabular}{lrrrrrlrrrrr}
\hline
         & \multicolumn{5}{c}{Singapore}                                                                                                                                                               &  & \multicolumn{5}{c}{NYC}                                                                                                                                                                \\ \cline{2-6} \cline{8-12} 
         & \multicolumn{1}{c}{Pt-Pl$\uparrow$} & \multicolumn{1}{c}{Pt-Pg$\uparrow$} & \multicolumn{1}{c}{Pl-Pl$\uparrow$} & \multicolumn{1}{c}{Pl-Pg$\uparrow$} & \multicolumn{1}{c}{Pg-Pg$\uparrow$} &  & \multicolumn{1}{c}{Pt-Pl$\uparrow$} & \multicolumn{1}{c}{Pt-Pg$\uparrow$} & \multicolumn{1}{c}{Pl-Pl$\uparrow$} & \multicolumn{1}{c}{Pl-Pg$\uparrow$} & \multicolumn{1}{c}{Pg-Pg$\uparrow$} \\ \hline
NUFTSPEC & \multicolumn{1}{c}{--}    & \multicolumn{1}{c}{--}    & \multicolumn{1}{c}{--}     & \multicolumn{1}{c}{--}     & 60.2$\pm$0.9                   &  & \multicolumn{1}{c}{--}    & \multicolumn{1}{c}{--}    & \multicolumn{1}{c}{--}     & \multicolumn{1}{c}{--}     & 58.5$\pm$0.8                   \\
T2VEC    & \multicolumn{1}{c}{--}    & \multicolumn{1}{c}{--}    & 72.8$\pm$2.3                   & \multicolumn{1}{c}{--}     & \multicolumn{1}{c}{--}     &  & \multicolumn{1}{c}{--}    & \multicolumn{1}{c}{--}    & 80.7$\pm$12.1                  & \multicolumn{1}{c}{--}     & \multicolumn{1}{c}{--}     \\
T-JEPA    & \multicolumn{1}{c}{--}    & \multicolumn{1}{c}{--}    & 75.4$\pm$1.8                   & \multicolumn{1}{c}{--}     & \multicolumn{1}{c}{--}     &  & \multicolumn{1}{c}{--}    & \multicolumn{1}{c}{--}    & 79.8$\pm$8.6                  & \multicolumn{1}{c}{--}     & \multicolumn{1}{c}{--}     \\
DIRECT   & 82.3$\pm$1.3                  & 84.3$\pm$0.5                  & 73.3$\pm$0.7                   & 36.8$\pm$1.0                   & 35.7$\pm$1.8                   &  & 84.6$\pm$1.1                  & 90.9$\pm$1.8                  & 74.5$\pm$0.8                   & 49.5$\pm$0.9                   & 44.6$\pm$2.3                   \\
TILE     & 79.0$\pm$2.1                  & 70.0$\pm$1.0                  & 50.5$\pm$0.5                   & 45.9$\pm$1.3                   & 41.1$\pm$1.3                   &  & 65.9$\pm$1.3                  & 78.3$\pm$0.7                  & 50.2$\pm$0.9                   & 49.4$\pm$3.8                   & 40.5$\pm$0.5                   \\
WRAP     & 88.6$\pm$0.3                  & 88.0$\pm$0.8                  & 71.6$\pm$1.1                   & 47.6$\pm$1.0                   & 47.6$\pm$1.0                   &  & 88.6$\pm$0.6                  & 88.0$\pm$1.7                  & 73.3$\pm$0.9                   & 55.0$\pm$1.1                   & 38.1$\pm$0.7                   \\
GRID     & 84.6$\pm$0.4                  & 84.4$\pm$0.4                  & 69.7$\pm$3.1                   & 45.8$\pm$0.4                   & 45.8$\pm$0.4                   &  & 82.2$\pm$3.9                  & 89.1$\pm$0.4                  & 73.9$\pm$0.9                   & 51.6$\pm$0.8                   & 38.1$\pm$3.1                   \\
THEORY   & 89.2$\pm$0.3                  & 90.0$\pm$0.5                  & 71.9$\pm$0.8                   & 45.0$\pm$1.0                   & 45.0$\pm$1.0                   &  & 89.7$\pm$0.8                  & 90.9$\pm$0.8                  & 73.4$\pm$0.8                   & 59.1$\pm$0.6                   & 45.5$\pm$4.1                   \\
Poly2Vec & 95.5$\pm$0.7                  & 94.9$\pm$0.2                  & 81.2$\pm$1.0                   & 50.9$\pm$0.8                   & 70.2$\pm$0.6                   &  & 95.3$\pm$0.3                  & 98.0$\pm$0.2                  & 83.0$\pm$0.4                   & 64.1$\pm$6.2                   & 68.4$\pm$0.8                   \\
Geo2Vec  & \textbf{98.5$\pm$0.3}                  & \textbf{96.1$\pm$0.2}                  & \textbf{96.4$\pm$0.5}                   & \textbf{61.2$\pm$0.4}                   & \textbf{75.6$\pm$0.4}                   &  & \textbf{98.7$\pm$0.4}                  & \textbf{99.1$\pm$0.3}                  & \textbf{98.9$\pm$0.3}                   & \textbf{67.5$\pm$0.8}                   & \textbf{70.0$\pm$0.4}                  \\ \hline
\end{tabular}
\end{table*}

We evaluated SRL methods based on their effectiveness in capturing shape and location. To further assess the quality of the learned representations, we tested them within a downstream GeoAI model. Details on the \textbf{experimental setup}, \textbf{hyperparameter sensitivity analysis}, \textbf{visualization results} and \textbf{performance discussions} are provided in the appendix.

\subsection{Datasets}
We used four datasets in our experiments: two with shape labels to evaluate the model’s performance on shape representation, and two real-world datasets to assess generalization in practical scenarios.\\
\textbf{MNIST}\cite{MNIST}: The original rasterized images are converted into polygon representations, and all digit shapes are randomly placed within a unit space. It contains 60,000 polygons, labeled according to its digit class. \\
\textbf{Building}\cite{Yan2021}: This dataset contains 5,000 building footprints, each manually labeled based on its geometric shape, where includes 10 common categories, such as E-shape, T-shape. \\
\textbf{Singapore}\cite{regiondcl}: This real-world dataset from OpenStreetMap includes 4,347 POIs, 45,634 roads and 109,877 buildings from the region of Singapore.  \\
\textbf{NYC}\cite{regiondcl}: Also sourced from OpenStreetMap, this dataset covers New York City and includes 14,943 POIs, 139,512 roads, and 1,153,008 buildings.

\subsection{Baselines}

Four types of baselines are included: \\
\textbf{Point encoders}: 
DIRECT~\cite{DIRECT}, directly utilizing coordinates; 
TILE~\cite{TILE}, partitions the whole area into tiles, and represents with tile embeddings; 
WARP~\cite{WARP}, uses a wrapping mechanism to encode points; 
GRID~\cite{NUFT}, multi-scale positional encoding based on Transformer's encoding; 
THEORY~\cite{SPACE2VEC}, encoding with unit vectors separated by 120\textdegree. \\
\textbf{Polyline encoders}: 
T2VEC~\cite{T2VEC}, GRU-based autoencoder to learn trajectory representations; 
T-JEPA~\cite{TJEPA}, contrastive learning-based trajectory representation learning method. \\
\textbf{Polygon encoders}: 
NUFTSPEC~\cite{NUFT}, encodes polygons through Fourier transform; 
PolyGNN~\cite{PolyGNN}, polygon encoder that encodes polygons and multi-polygons with GNN. \\
\textbf{Unified encoder}: 
Poly2Vec~\cite{poly2vec}, decomposes points, polylines, and polygons, and encodes them by geometrically sampling from the Fourier spectral space.

\subsection{Effectiveness of Shape Representation}

\begin{table*}[]
\centering
\setlength{\tabcolsep}{2pt} 
\caption{Comparison of spatial representation learning methods on Land Use Classification and Population Prediction tasks.}
\label{tab:5}
\begin{tabular}{lccclccc}
\hline
\multicolumn{1}{c}{} & \multicolumn{7}{c}{Land Use Classification}                                                                                                                                         \\ \cline{2-8} 
                     & \multicolumn{3}{c}{Singapore}                                            &                               & \multicolumn{3}{c}{NYC}                                                  \\ \cline{2-4} \cline{6-8} 
                     & L1$\downarrow$                      & KL$\downarrow$                      & Cosine$\uparrow$               &                               & L1$\downarrow$                      & KL$\downarrow$                      & Cosine$\uparrow$               \\ \cline{1-4} \cline{6-8} 
RegionDCL            & 0.498$\pm$0.038             & 0.294$\pm$0.047             & 0.879$\pm$0.021          &                               & 0.418$\pm$0.012             & 0.229$\pm$0.013             & 0.912$\pm$0.006          \\
RegionDCL w/ Poly2Vec             & 0.484$\pm$0.021             & \textbf{0.278$\pm$0.025}             & 0.881$\pm$0.012          &                               & 0.397$\pm$0.010             & 0.212$\pm$0.011             & 0.923$\pm$0.007          \\
RegionDCL w/o Rotation             & 0.493$\pm$0.054             & 0.309$\pm$0.068             & 0.872$\pm$0.028          &                               & 0.408$\pm$0.014             & 0.226$\pm$0.021             & 0.913$\pm$0.008          \\
RegionDCL w/ Geo2Vec              & \textbf{0.475$\pm$0.053}    & 0.287$\pm$0.058    & \textbf{0.884$\pm$0.025} & \textbf{}                     & \textbf{0.390$\pm$0.013}    & \textbf{0.208$\pm$0.017}    & \textbf{0.928$\pm$0.007} \\ \hline
                     & \multicolumn{7}{c}{Population Prediction}                                                                                                                                           \\ \cline{2-8} 
                     & \multicolumn{3}{c}{Singapore}                                            &                               & \multicolumn{3}{c}{NYC}                                                  \\ \cline{2-4} \cline{6-8} 
                     & MAE$\downarrow$                     & RMSE$\downarrow$                    & R$^2\uparrow$                    &                               & MAE$\downarrow$                     & RMSE$\downarrow$                    & R$^2\uparrow$                    \\ \cline{1-4} \cline{6-8} 
RegionDCL            & 5807.54$\pm$522.74          & 7942.74$\pm$779.44          & 0.427$\pm$0.108          & \multicolumn{1}{c}{}          & 5020.20$\pm$216.63          & 6960.51$\pm$282.35          & 0.575$\pm$0.039          \\
RegionDCL w/ Poly2Vec             & 4957.58$\pm$506.02          & 6874.47$\pm$851.73          & 0.561$\pm$0.117          & \multicolumn{1}{c}{}          & 4602.75$\pm$179.66          & 6393.38$\pm$279.70          & 0.621$\pm$0.037          \\
RegionDCL w/ Geo2Vec              & \textbf{4658.51$\pm$483.02} & \textbf{6515.26$\pm$795.91} & \textbf{0.585$\pm$0.156} & \multicolumn{1}{c}{\textbf{}} & \textbf{4486.49$\pm$163.65} & \textbf{6189.85+280.05} & \textbf{0.625$\pm$0.055} \\ \hline
\end{tabular}
\end{table*}

We evaluate the effectiveness of our polygon shape representation through two tasks: \textbf{shape classification} (Shape) and \textbf{predicting the number of edges} (Edge), reporting accuracy and Mean Absolute Error (MAE), respectively. As depicted in Table \ref{Table 1}, Geo2Vec significantly outperforms all baselines. PolyGNN, as a GNN-based method, shows poor performance when modeling complex polygons from the MNIST dataset. Moreover, the method relies on contrastive learning and is not able to learn general-purpose polygon representations, limiting its applicability to regression tasks.

To evaluate the model's performance on line entities, we use the learned representations to infer the \textbf{length of lines} in two real-world datasets. Results in Table \ref{Table 2} show the superior performance of Geo2Vec (almost 2$\times$ improvement over the best baseline). RNN-based approaches like T2Vec and T-JEPA primarily model relationships between individual vertices, overlooking line segments and thereby limiting their ability to represent line entities effectively.

Additionally, we observe that the embeddings learned by Geo2Vec consistently exhibit the lowest standard deviation across nearly all tasks, which holds throughout almost all our experiments. This indicates that the learned embedding space is well-structured and robust.

\subsection{Effectiveness of Location Representation}

To evaluate the effectiveness of location representations generated by different methods, we employ two basic spatial reasoning tasks: \textbf{distance estimation} (Table \ref{Table 3}) and \textbf{topological relationship classification} (Table \ref{Table 4}). We report MAE and accuracy for evaluation. To assess the uniformity of the learned representations, we measure their performance when inferring across different types of geo-entities. For brevity, we denote Point as \textbf{Pt}, Polyline as \textbf{Pl}, and Polygon as \textbf{Pg} in the following two tables.

Geo2Vec consistently outperforms all baselines across various distance estimation scenarios. In particular, for complex distance pairs such as polygon-to-polygon and polygon-to-polyline, the performance improvement over the SOTA methods is at least 54.3$\%$. 

For topological relationship classification, Pt-Pl, Pt-Pg, and Pl-Pl are binary tasks, while Pl-Pg and Pg-Pg are multiclass. Details can be found in the Appendix. Geo2Vec outperforms both specialized encoders and the unified encoder. The most significant improvement is observed in Pl–Pl relationship inference, where Geo2Vec achieves at least an 18.7$\%$ increase in accuracy. 

\subsection{Effectiveness in GeoAI Model}

We further evaluate representation effectiveness within existing GeoAI models. This experiment shows the practical effectiveness and real-world potential of Geo2Vec.

Following the experimental setup of previous work \cite{poly2vec}, we adopt \textbf{RegionDCL} \cite{regiondcl} as our GeoAI pipeline model. RegionDCL is designed to learn region-level representations based on the spatial distribution and shape of buildings. The effectiveness of the learned representations is evaluated through two downstream tasks: \textbf{Land Use Classification} and \textbf{Population Prediction}. In the original setting, each building is rasterized into an image, and its representation is extracted using a Convolutional Neural Network. Since rasterization discards location information, the model incorporates a distance-biased Transformer to reintroduce spatial relationships. In our experiment, we modify this pipeline by replacing it with a standard Transformer network. Instead of using CNN-extracted features, we directly input features obtained from Geo2Vec and Poly2Vec.

In Table \ref{tab:5}, \textbf{RegionDCL w/o Rotation} refers to the Geo2Vec model without rotation-invariant positional encoding, while \textbf{RegionDCL w/ Geo2Vec} represents the full Geo2Vec model. The representations generated by Geo2Vec enable RegionDCL to produce the highest-quality region embeddings. We attribute this improvement to the learning-friendly shape information, and global location information captured by Geo2Vec, which is absent in the raster representation of buildings.\footnote{See Appendix for an explanation of the limited improvement, due to RegionDCL’s performance ceiling with this input type.} The ablation experiment shows that the rotation-invariant property preserved by our positional encoding is beneficial for unsupervised downstream GeoAI model like RegionDCL.

\subsection{Efficiency Analysis}
\begin{figure}[h]
    \centering
    \begin{subfigure}[b]{0.48\columnwidth}
        \includegraphics[width=\linewidth]{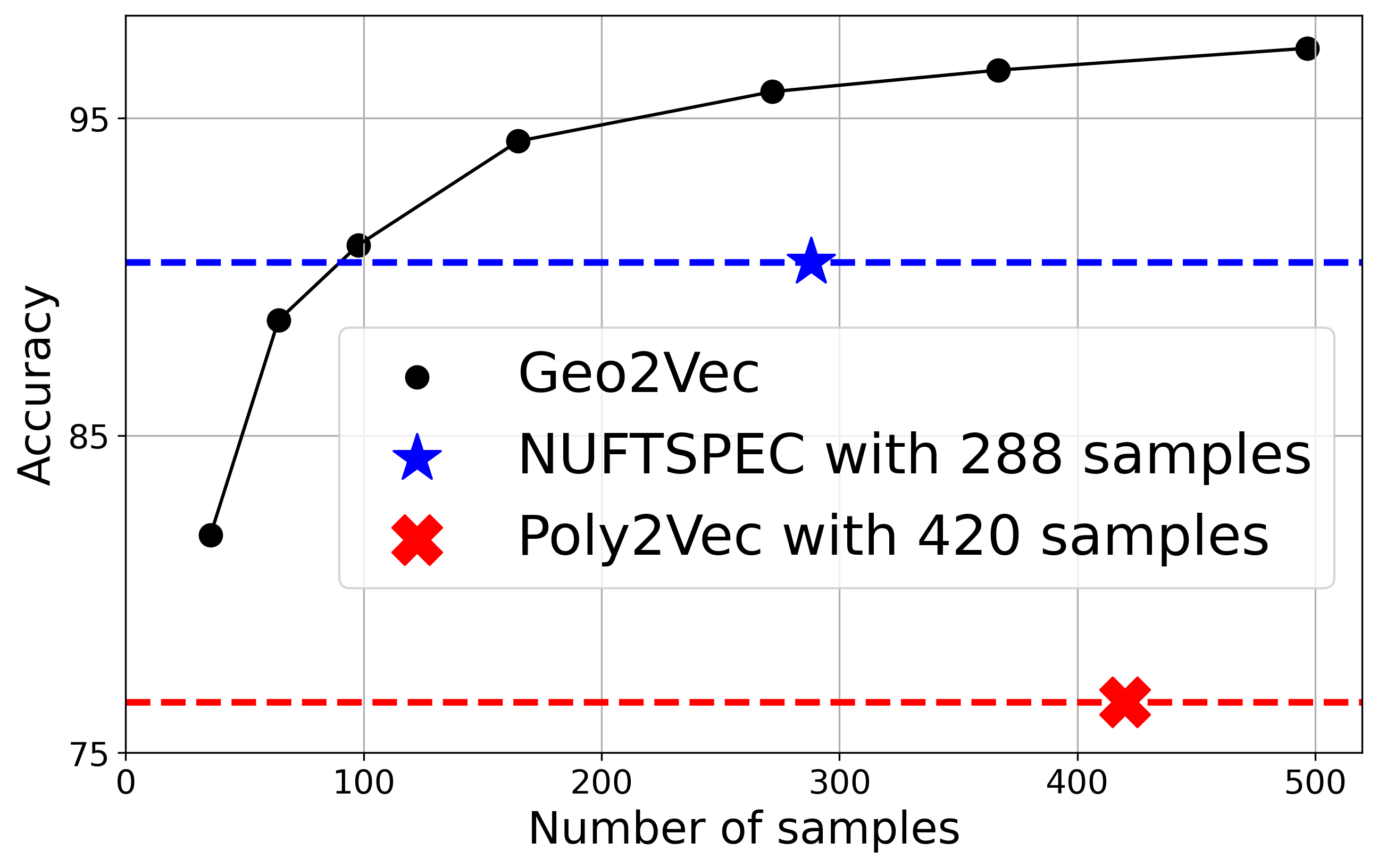}
        \caption{Classification-Accuracy \(\uparrow\)}
        \label{fig:sub1}
    \end{subfigure}
    \hfill
    \begin{subfigure}[b]{0.48\columnwidth}
        \includegraphics[width=\linewidth]{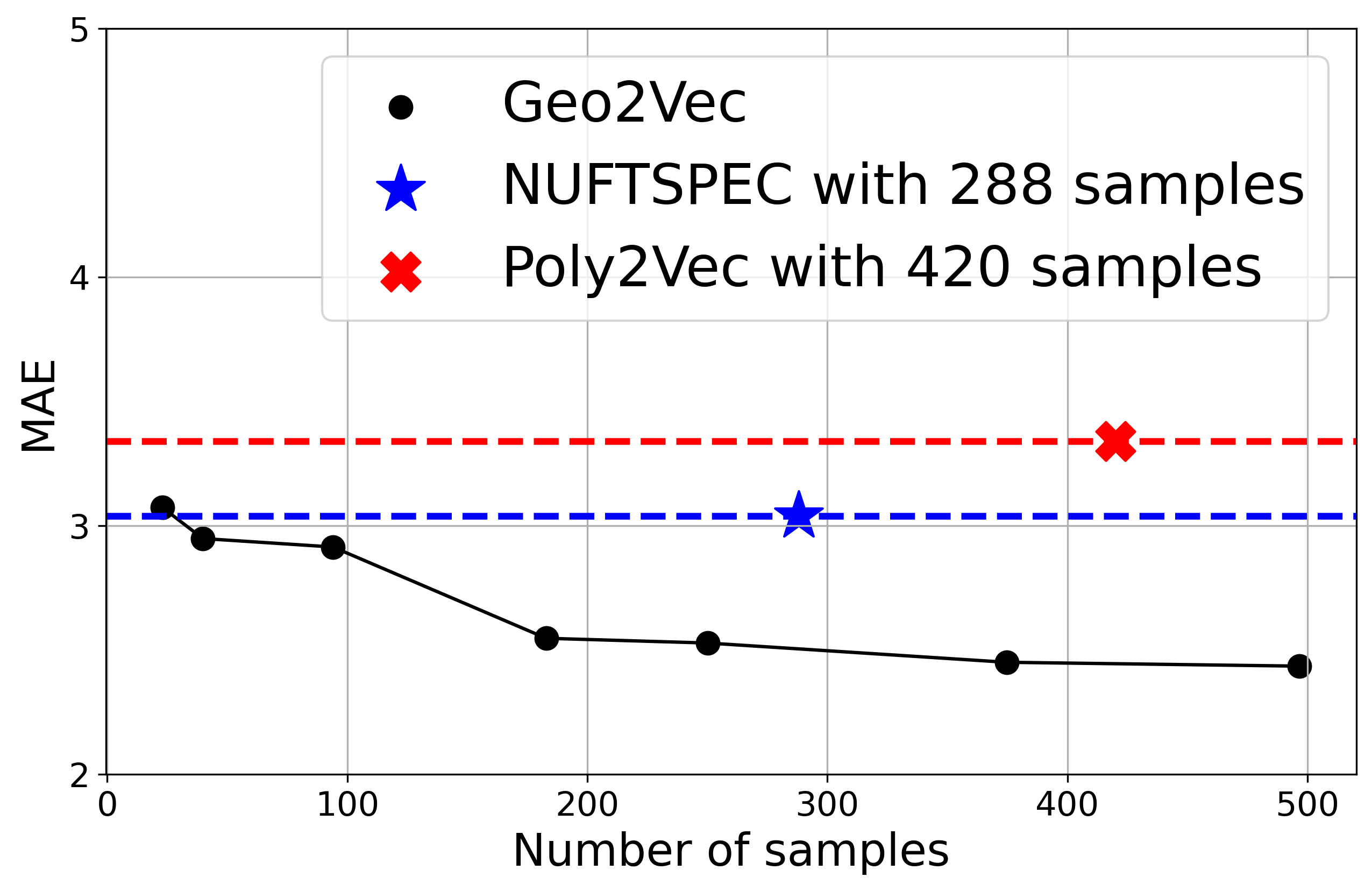}
        \caption{Edge-MAE \(\downarrow\)}
        \label{fig:sub2}
    \end{subfigure}
    \caption{Comparison between number of sampled points and models' performance on the \textbf{Building} dataset.}
    \label{Figure 3}
\end{figure}
Previous experiments have showed that Geo2Vec achieves superior embedding quality compared to existing methods under the same embedding dimensionality. We now compare their performance in terms of the number of sample points required. As shown in Figure~\ref{Figure 3}, spectral methods such as NUFTSPEC and Poly2Vec rely on sampling in the Fourier domain, using 288 and 420 points, respectively. However, benefiting from direct access to coordinate space and adaptive sampling, Geo2Vec requires significantly fewer sample points to achieve the same performance. Highlighting the effectiveness of learning geo-entity representations directly from coordinate space rather than relying on less interpretable spectral features.

\section{Conclusion}
In this paper, we proposed a unified spatial representation learning method, which is generalizable to all types of geo-entities, including multipolygons and polygons with holes. The learned spatial representation shows superior performance on tasks such as shape classification, distance estimation, and topological relationship classification. Through experiments with an existing GeoAI model, we further show its practicality in real-world scenarios.

To the best of our knowledge, this is the first study to learn geo-entity representations directly from coordinate space, without relying on decomposition or Fourier transform techniques. Our research reveals the possibility of using neural networks to directly learn both the location and shape representations of geo-entities, and serves as a promising step toward the development of future representation methods for geo-entities.

\bibliography{aaai25}

\clearpage


\section*{Supplementary material}

# Appendix

## Experiment Setting

For polygon shape learning, we scale each polygon individually to a canonical space $[-1, 1] \times [-1, 1]$ and then learn its shape embedding. For location representation learning, we normalize the entire dataset to a canonical space before applying the learning algorithm. All distance- and length-related results reported in the paper are computed within this canonical coordinate system.

For all the shape representation testing tasks, we split each dataset into training, validation, and test sets using a 70:15:15 ratio. A Multi-Layer Perceptron (MLP) is trained on the 70% training dataset, and then testing and evaluating on the rest two datasets. And for all the experiments, we randomly split the dataset and train the MLP for 5 times and report average performances and standard deviation.

For distance estimation and topological relationship classification, we follow the experiment settings of previous research, details can be found at (Siampou et al. 2025).

To ensure fair comparison across methods, we fix the output dimensionality of all representation learning approaches: 256 dimensions for shape and 256 dimensions for location. Aside from this adjustment, all other parameters follow the original settings specified in the respective papers.

For Geo2Vec, Sample Density $\epsilon$ is set as 100 for shape learning and 50000 for location learning, Uniform Sample Density $N_{\text{axis}}$ is set as 10 for shape learning and 30 for location learning, Batch Size $b$ is set as $1024 \times 20$. We set the frequency range as $L_{\min} = \log_2\left(\frac{1}{\pi}\right), L_{\max} = \log_2\left(\frac{6}{\pi}\right)$ for location representation, and $L_{\min} = 0, L_{\max} = 8$ for shape representation. For shape representation learning, we set $L = 8$, for location representation leaning, we set $L = 16$, meaning we uniformly sample 8 and 16 frequency bands on a logarithmic scale between $L_{\min}$ and $L_{\max}$.

The experiment is conducted on a cluster node equipped with an 16-core CPU, 64GB of memory, and one 12GB NVIDIA GeForce RTX 3060 GPU.

## Topological Relationship Classification

We evaluated the embedding's capability in topological relationship classification, where relationships between geo-entity pairs are defined by the DE-9IM model (Clementini, Di Felice, and van Oosterom 1993). For clarity, we denote Point as **Pt**, Polyline as **Pl**, and Polygon as **Pg** in the following two tables. Details are shown in Table 1.

Table 1: Topological relationships of geo-entity pairs.

| Geo-entity Pair | Topological Relationships (a relationship b) |
|---|---|
| Pt-Pl | disjoint, intersects |
| Pt-Pg | disjoint, contains |
| Pl-Pl | disjoint, intersects |
| Pl-Pg | disjoint, touches, intersects, within |
| Pg-Pg | disjiont, touches, intersects, contains, within, equals |

## Parameter Sensitivity

Firstly, we analyze the effect of varying the number of layers. On the **Building** dataset, we evaluate model performance using SDF test MAE and number of edge prediction MAE. Lower test SDF-MAE means model learns SDFs better. As shown in Figure 1, increasing the number of layers improves the model's ability to learn the SDF, resulting in lower SDF prediction error. This, in turn, enhances the quality of the learned shape features, leading to more accurate number of edge predictions and reduced variance. To balance computational efficiency and performance, we chose an 8-layer architecture as the default.

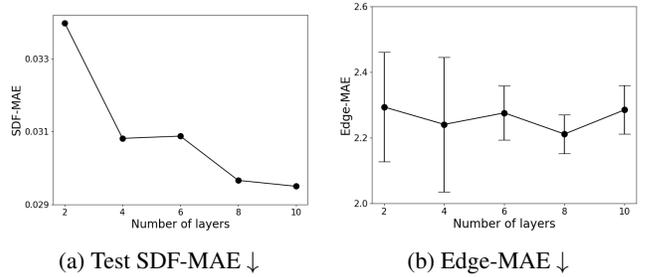

(a) Test SDF-MAE ↓    (b) Edge-MAE ↓

Figure 1: Effect of number of layers in Geo2Vec to model's performance on the **Building** dataset.

Besides the number of layers, we also evaluated the model's performance under different settings of $L$, the number of sampling frequencies used in the Positional Encoding function. To investigate how positional encoding facilitates learning fine-grained characteristics of geo-entities, we conducted experiments on the **MNIST** dataset, which contains more complex polygonal shapes than the other datasets. The results are shown in Figure 2. As the number of frequencies increases, the period of the positional encoding function decreases, producing higher-frequency features that better capture fine-scale variations in the SDF. However, these strong high frequency signals also reduce the model's ability to learn more generalizable patterns. This explains why increasing the number of frequencies improves performance on edge count prediction, but degrades the quality of SDF simulation. To balance the capture of fine-grained features and the overall quality of the learned SDF, we set $L = 8$ as the default configuration in our experiments.

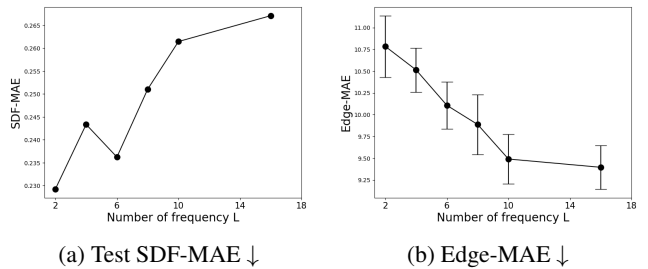

(a) Test SDF-MAE ↓    (b) Edge-MAE ↓

Figure 2: Effect of number of number of frequencies $L$ in Positional Encoding of Geo2Vec on the **MNIST** dataset.

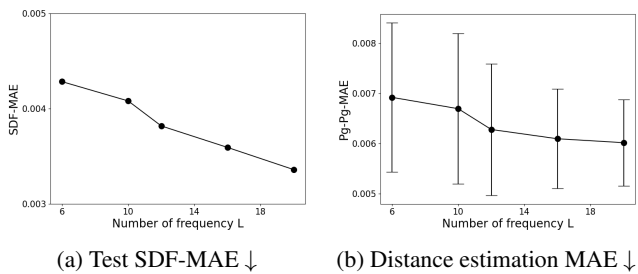

(a) Test SDF-MAE ↓    (b) Distance estimation MAE ↓

Figure 3: Effect of number of number of frequencies $L$ in Positional Encoding of Geo2Vec on the location learning capability on the **Building** dataset.

We analyze the effect of positional encoding on location representation learning. As shown in Figure 3, experiments on the **Building** dataset demonstrate that increasing the number of frequencies $L$ in the positional encoding improves the model's ability to capture the overall variation patterns of the SDF. This is further supported by improved performance on the distance estimation task, indicating that better learning of SDFs enhances the model's ability to represent location information and thus benefits downstream tasks. We set $L = 16$ in our experiment.

**Visualization of SDFs**

In Figure 4, we visualize several representative samples of the learned SDFs and their corresponding ground truth for both digits and buildings. As shown in the figure, Geo2Vec effectively captures fine-grained details of complex real-world polygons from **Singapore** dataset, such as building footprints, and also models smooth shape variations, particularly for curved structures like the digits in the **MNIST** dataset.

**Performance Discussion**

In our experiment, Geo2Vec shows superior performance over SOTA on all spatial reasoning tasks, while the improvement on the RegionDCL is marginal. This is likely because RegionDCL is already near its performance ceiling. RegionDCL is an unsupervised model and is not learning end-to-end, which means, it cannot fully exploit the richer features Geo2Vec provides. Additionally, relying solely on building data limits performance in tasks like land use classification and population prediction.

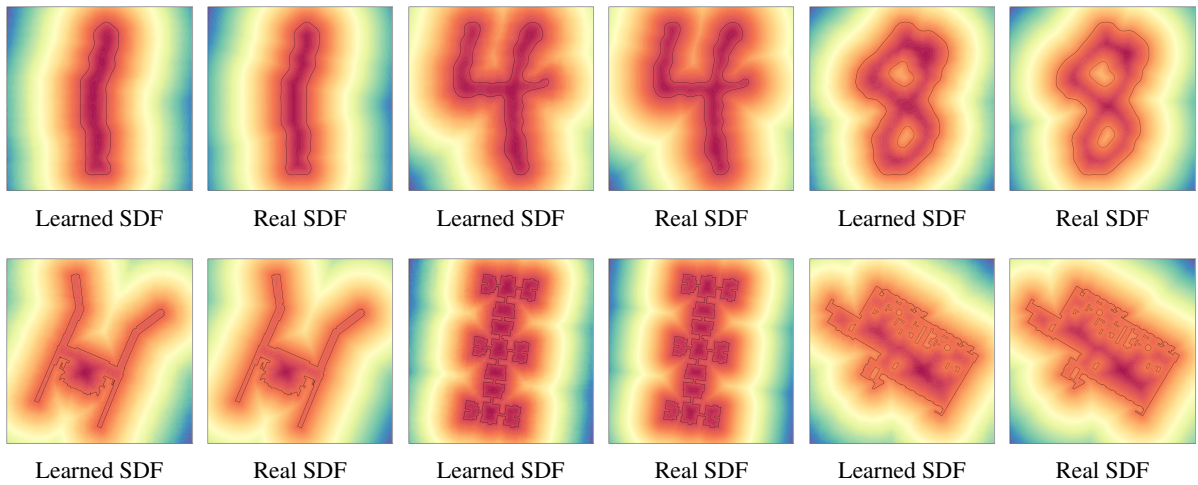

Figure 4: Visualization of the learned SDFs by Geo2Vec and the ground truth. The top row shows digits from the **MNIST** dataset, while the bottom row displays real-world buildings from the **Singapore** dataset.

 

\end{document}